\documentclass{article}
\usepackage{spconf,amsmath,graphicx,hyperref}

\usepackage{array}
\usepackage[table]{xcolor}
\usepackage{booktabs}

\title{Towards Orthographically-Informed Evaluation of \\
Speech Recognition Systems for Indian Languages}
\name{\begin{tabular}{c}
Kaushal Santosh Bhogale$^{\star1}$, Tahir Javed$^{\star1}$, Greeshma Susan John$^{\star1}$,\\
Dhruv Rathi$^{2}$, Akshayasree Padmanaban$^{1}$, Niharika Parasa$^{1}$, Mitesh M. Khapra$^{1}$
\end{tabular}}

\address{\begin{tabular}{c}
$^{1}$AI4Bharat, WSAI, Indian Institute of Technology Madras, India, $^{2}$Sarvam AI, India\\
\texttt{cs22d006@cse.iitm.ac.in, miteshk@dsai.iitm.ac.in}
\end{tabular}}
      
\begin{document}
\ninept
\maketitle
\begin{abstract}
Evaluating ASR systems for Indian languages is challenging due to spelling variations, suffix splitting flexibility, and non-standard spellings in code-mixed words. Traditional Word Error Rate (WER) often presents a bleaker picture of system performance than what human users perceive. Better aligning evaluation with real-world performance requires capturing permissible orthographic variations, which is extremely challenging for under-resourced Indian languages. Leveraging recent advances in LLMs, we propose a framework for creating benchmarks that capture permissible variations. Through extensive experiments, we demonstrate that OIWER, by accounting for orthographic variations, reduces pessimistic error rates (an average improvement of 6.3 points), narrows inflated model gaps (e.g., Gemini–Canary performance difference drops from 18.1 to 11.5 points), and aligns more closely with human perception than prior methods like WER-SN by 4.9 points.
\end{abstract}
\section{Introduction}
\label{sec:intro}

Over the past few years, steady progress has been made in creating datasets \cite{kathbath, indicvoices} and building models \cite{indicwav2vec, vistaar, indicconformer} for automatic speech recognition (ASR) in Indian languages, aiming to bring performance closer to that of high-resource languages such as English. Despite these advances, the word error rate (WER) for Indian languages has remained significantly higher, often in stark contrast to the single-digit WER reported for English systems \cite{open-asr-leaderboard}.

A closer examination reveals that much of this gap arises not solely from ASR model errors but from inflated WER values caused by the absence of well-established orthographic standards and consistent transcription guidelines for Indian languages. This issue stems from the inherent linguistic richness of these languages: flexible spelling variations for common suffixes, permissibility in splitting or merging morphemes, and the prevalence of non-standard spellings in code-mixed contexts. Such code-mixed terms, although widespread in everyday speech, have not yet been fully standardized. Existing transcription guidelines only partially capture these permissible variations, leaving many uncodified. Consequently, stylistic inconsistencies in transcriptions introduce noise into evaluations, causing ASR systems to appear less accurate than they are. 

Prior work has attempted to address this challenge through multi-reference WER \cite{arabic2015, arabic2019, africanenglish, style_agnostic}, wherein multiple independent transcriptions are collected for the same audio. ASR predictions are scored against each reference, with the minimum WER taken as the final score. While effective in improving robustness, this approach is resource-intensive, demanding substantial human effort and cost. 
Furthermore, a small set of references may still fail to capture the full space of valid orthographic variants. An alternative strategy is to ensure comprehensive coverage of permissible variations (see Table \ref{tab:example}). Prior attempts in this direction have relied on manual curation \cite{sclite}, extraction from social media sources such as Twitter \cite{werd}, or heuristic normalization rules \cite{snwer}. However, these methods often demand substantial annotation effort, access to large text corpora, or specialized pipelines, all of which are limited for low-resource Indian languages.

In this paper, we present a framework for creating orthographically informed benchmarks that systematically capture possible spelling and word-formation variations. Our method augments existing benchmarks by generating exhaustive lists of candidate variants for each word. While manually compiling such lists is infeasible, we leverage large language models (LLMs) to automatically propose variations, requiring human input only for refinement and correction.
To integrate these word-level variations into evaluation, we introduce the Orthographically Informed Word Error Rate (OIWER) metric. Through extensive experiments, we demonstrate that by accounting for orthographic variations, OIWER reduces pessimistic error rates, improving scores by 6.3 points on average across all languages for the state of the art Canary model \cite{canary}, and yields error estimates that more closely align with human perception. Further, OIWER narrows exaggerated performance gaps between models (e.g., differences between the Gemini model and the Canary model drop from 18.1 to 11.5 points on average), providing a truer sense of progress. Lastly, compared to prior normalization-based methods like WER-SN \cite{snwer}, OIWER yields better alignment with human perception than prior methods like WER-SN by 4.9 points. The code is available at this \href{https://github.com/AI4Bharat/OIWER-Orthographically-Informed-Benchmarking-for-ASR/}{GitHub} page.

\section{Related Work}

To account for spelling variations and dialectal differences, \cite{arabic2015} introduced multi-reference WER (MR-WER) for dialectal Arabic ASR, which was later extended to other dialects \cite{arabic2019}. Similarly, \cite{japanese} proposed ``lenient'' character error rate (CER) for Japanese using lattice-based references, while \cite{style_agnostic} applied multi-reference schemes for English stylistic variations. However, collecting multiple transcriptions remains costly.
Manual rule-based methods such as SCLITE \cite{sclite} support Global Mapping files for specifying permissible substitutions. However, defining such mappings is particularly challenging for Indian languages, which exhibit substantial orthographic variation. Many code-mixed words show five or more spellings, making exhaustive enumeration impractical. WERd \cite{werd} automated this by mining variants from social media corpora, but such resources are scarce for most Indian languages. 

In the Indian context, \cite{malayalam_complex} showed WER over-penalizes valid orthographic variations in morphologically rich languages like Malayalam. Text normalization \cite{whisper} can standardize ASR outputs; however, \cite{malayalam_normalization} report that such methods often distort transcriptions by removing critical vowel signs, resulting in misleadingly low WER scores. As an alternative, \cite{malayalam_cer} proposed using Character Error Rate (CER); however, this metric may overestimate performance, since even a single character error can necessitate retyping entire words on Indic keyboards.
Similarly, poWER \cite{power} introduced a framework for code-switched homophones using WX-based pronunciation schemes, while \cite{snwer} generated variants through spelling and segmentation normalization for four Indian languages. Both approaches rely heavily on auxiliary resources like grapheme-to-phoneme converters and transliteration tools. For low-resource Indian languages, such resources are often incomplete or inconsistent. 

\section{Inflated WER in Indian Languages}
\label{sec:motivation}
\begin{figure}
\centering
\includegraphics[width=0.8\linewidth]{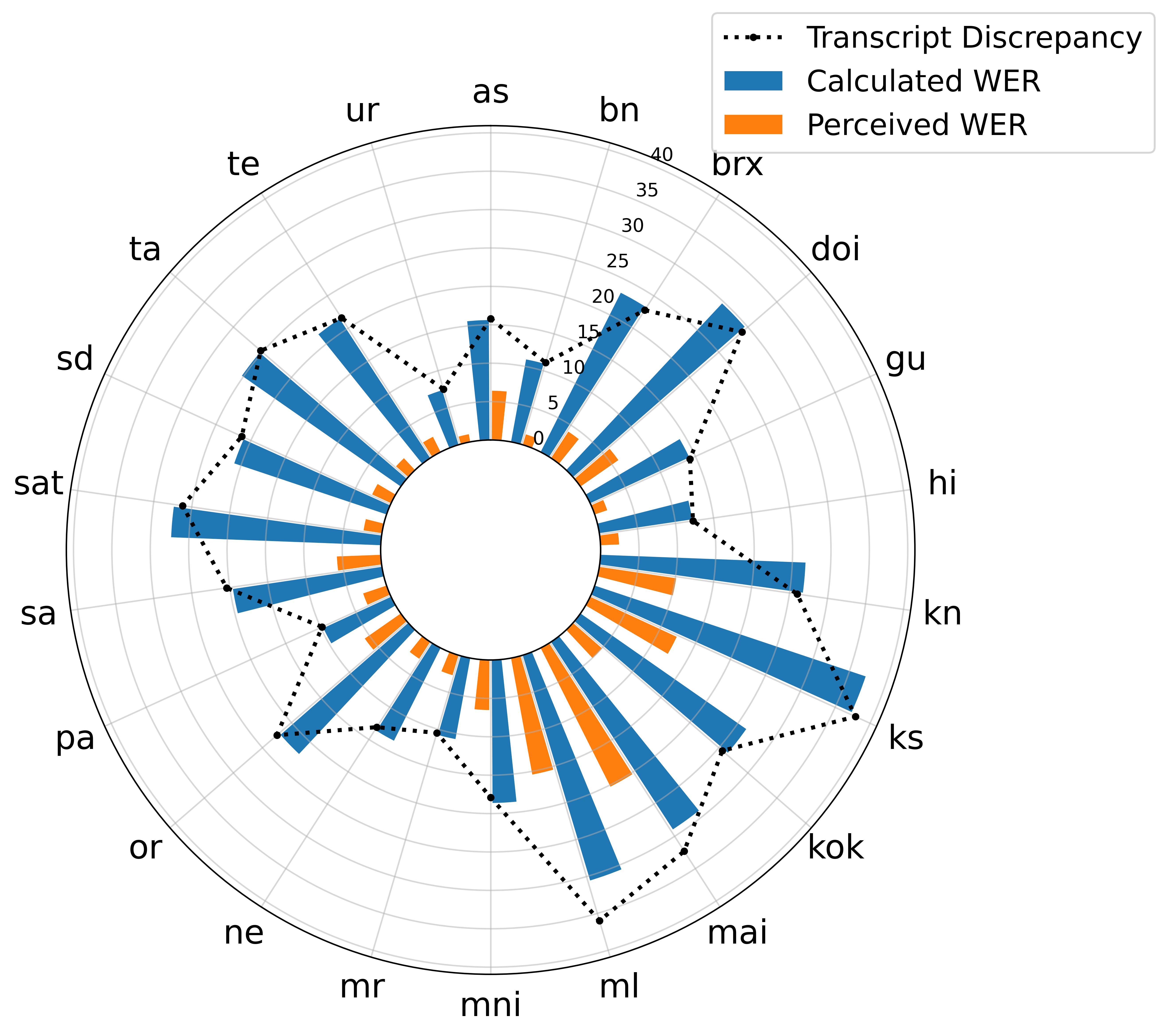}
\caption{For Indian languages, WER reflects inflated values, much higher than perceived error (orange and blue). Moreover, there is a large discrepancy between two valid transcripts (black).}
\label{fig:motivation}
\end{figure}

Before proposing a solution, we first characterize the issue of inflated WERs due to orthographic variations. To establish its extent and impact, we evaluate an existing model to demonstrate that the problem of accounting for orthographic variations is real and significant. To this end, we evaluated a Canary \cite{canary} model fine-tuned on the MahaDhwani \cite{indicconformer} dataset, using the IndicVoices \cite{indicvoices} benchmark spanning 22 Indian languages across four diverse language families.
In our initial evaluation, WER was computed against the original human-generated transcripts, yielding consistently high error rates (blue in Figure \ref{fig:motivation}). We then asked experienced transcribers to post-edit the model outputs, treating these human-corrected transcripts as an alternative gold standard. When recalculated against this new reference, WER dropped substantially (orange in Figure \ref{fig:motivation}), since the transcribers retained valid orthographic variations present in the model predictions. Ideally, faithful reconstruction of the original transcript would result in zero WER between the two human references, i.e., the original one and the post-edited one. However, when averaged across all languages, the discrepancy between the original and post-edited transcripts was 20.4 WER (black in Figure \ref{fig:motivation}).

Based on our analysis of this data, we find that there are three main factors behind these pessimistic WER values: (1) permissible spelling variations between predicted and reference transcripts, (2) human transcription errors in a few cases, and (3) genuine audio ambiguities where, beyond spelling differences, several valid interpretations may exist in the absence of further context. In this work, we focus on the first factor and propose an approach to account for orthographic variations while computing Word Error Rates.

\section{A framework to create Orthographically Informed (OI) benchmarks}

In this section, we propose a schema for creating orthographically informed benchmarks that systematically capture spelling variations. 

\subsection{Identifying types of variations} As a first step, we adopt a linguistically \label{sec:variations}
grounded approach to identify the different types of orthographic variations possible in Indian languages. We do so by consulting language experts from each language and identify the categories of variations listed below. We then come up with a human-in-the-loop approach to generate variations. 

\noindent \textbf{Matra and Diacritic Variations.} Matras are diacritical marks or vowel signs that modify consonants to represent different vowel sounds, and are used in all Indian scripts. For robust evaluations, we consider all alternative forms of vowels and diacritical marks including \textit{nuktas} (a diacritical dot placed below certain consonants) in Hindi and \textit{pulli} (to indicate pure consonants) in Tamil.

\noindent \textbf{Spelling Variations for Loaned Words.} Several Indian languages have loaned words from English, Sanskrit, Persian or other languages that often have multiple recognized spellings. We consider all plausible forms of such words, including transliterations and localized versions.

\noindent \textbf{Splitting / Merging Compound Words.} We need to consider whether a compound word can be decomposed into its constituent parts, and whether two separate words can be combined to form a single word as a valid variant. This is a common variation type for agglutinative languages like Malayalam.

\noindent \textbf{Phonetic Variations.} We consider spelling variations due to regional accents, dialects or pronunciations.

\noindent \textbf{Ligature Variations.} Some Indic scripts allow consecutive consonants to be joined into a ligature or remain separate. We consider variations both with and without ligatures.

\noindent \textbf{Sandhi.} Sandhi rules govern the merging and transformation of sounds at word or morpheme boundaries in Indian languages, altering letters based on phonetic conditions. These rules help in splitting complex words into meaningful components and merging smaller units into one. 
We consider all variations possible by applying sandhi rules to words.

\noindent \textbf{Inverse Text Normalization (ITN).} Spoken utterances involving numbers, dates, currencies, units, and other semiotic tokens can be validly represented in multiple orthographic forms. We account for all such permissible ITN variations.


\begin{table}[]
    \centering
    \footnotesize
    \caption{An example of the original transcript ($T$) and the set of variations ($V$) generated for different words (indicated in \textcolor{blue}{blue}). While our study focuses on Indian languages, we include an English example to enhance readability for a broader audience.}
    \renewcommand{\arraystretch}{1.1}
    \begin{tabular}{m{0.2cm}|m{6.7cm}}
         \multicolumn{1}{c|}{$T$} &  \emph{``Can I find the instructions to download the EPFO passbook statement for my account number five six eight four nine, 56849] in the blue coloured catalogue?''} \\\hline
         \multicolumn{1}{c|}{$V$} & {``Can I find the instructions to download the \textcolor{blue}{[EPFO, E P F O]} \textcolor{blue}{[passbook, pass book]} statement for my account number \textcolor{blue}{[five six eight four nine, 56849]} in the blue \textcolor{blue}{[coloured, colored]} \textcolor{blue}{[catalogue, catalog]}''}
    \end{tabular}
    \label{tab:example}
\end{table}

\subsection{LLM-assisted generation of word variations}

In many cases, the possible variations of a word depend on the context of the sentence, particularly in rules governing word merging and splitting. Therefore, instead of generating variations for words in isolation, we take the entire transcript of the audio into account to generate context-aware variations.

\noindent\textbf{Objective.} Given an audio transcript \( T = (T_1, T_2, \dots, T_K) \), which represents an ordered sequence of \( K \) words, the objective is to generate a set of variations \( V \) for a given utterance \( T \). The variations \( V \) are represented as an ordered sequence of \( L \) sets, where each set $V_l$ corresponds to a set of variations for a specific subsequence of words 
\( (T_s, \dots, T_e) \) where $s$ and $e$ are the start and end index of the subsequence in $T$. We allow for a subsequence of words to be considered in a single variation to account for merging of words. Note that, by design $L \leq K$. We show an example of $T$ and $V$ in Table \ref{tab:example}.

\noindent\textbf{Generation.} We prompt a text LLM to generate variations for all words in the transcript $T$. To ensure consistency, we provide the LLM with predefined categories of variations in Indian languages as listed in section \ref{sec:variations}. We also provide representative examples for each category as guidelines.
Additionally, we specify the output format for $V$ and instruct the LLM to generate variations as a sequence of sets, where the variations for each word are separated by commas. 

\noindent\textbf{Post-editing.} We request human experts to review the generated variations, allowing annotators to refine the sets corresponding to each word. They can add missing variants or remove incorrectly generated ones to ensure accuracy. We developed a custom web interface using LabelStudio \cite{labelstudio} to enable efficient human review, allowing users to add or remove elements with minimal effort.

\subsection{Orthographically-Informed Word Error Rate}

We modify the Word Error Rate algorithm to account for word variations. WER is calculated by computing the minimum edit distance between ground truth $T$ and the model prediction \( P = (P_1, P_2, . . . , P_N) \).
We calculate minimum edit distance between the predicted transcript ($P$) and reference variations ($V$) by comparing predicted words with each possible variation in the reference.
Following the SCLITE toolkit \cite{sclite}, we implement this using Dynamic Programming (DP). The DP table records the minimum edit distance to align the first $k$ segments of variations $V$ with the first $j$ words of prediction $P$.
By exploring all alignments through DP recurrence, the algorithm identifies optimal matching paths between $P$ and $V$. We term this metric Orthographically-Informed WER (OIWER).

\section{Experimental Setup}


\noindent\textbf{Creation of Orthographically-Informed IndicVoices benchmark.} We applied our framework to the IndicVoices \cite{indicvoices} benchmark using Gemini-2.5-Pro to generate word variations. We recruited 61 native-speaking transcribers from the 22 languages, providing them with guidelines and 5 example variations for each type. Table \ref{tab:dataset_stats} presents statistics on utterances, words, and variations created. The number of variations per word ranges from 1.3 to 3.2 across languages, with higher values for low-resource languages like Kashmiri. Moreover, post-editing orthographic variations requires only 1.2 minutes per utterance compared to 2.4 minutes for manual transcription, making our approach both more time-efficient and more comprehensive than multi-reference WER methods.

\begingroup
\setlength{\tabcolsep}{2.5pt} 
\begin{table}[]
\centering
\caption{Statistics of the Orthographically-Informed IndicVoices Benchmark (utts: utterances, vars: variations)}
\label{tab:dataset_stats}
\small

\begin{tabular}{l|ccc||l|ccc}
\toprule
\textbf{lang} & \textbf{\# utts} & \textbf{\# words} & \textbf{\# vars} & \textbf{lang} & \textbf{\# utts} & \textbf{\# words} & \textbf{\# vars} \\
\midrule
\textbf{as} & 3.0K & 40.2K & 89.1K & \textbf{mni} & 1.7K & 15.1K & 27.5K \\
\textbf{bn} & 2.2K & 30.2K & 61.0K & \textbf{mr} & 2.3K & 32.3K & 76.0K \\
\textbf{brx} & 1.8K & 17.5K & 28.5K & \textbf{ne} & 2.5K & 33.4K & 73.3K \\
\textbf{doi} & 0.5K & 6.9K & 18.7K & \textbf{or} & 1.6K & 19.6K & 50.9K \\
\textbf{gu} & 2.2K & 34.0K & 66.4K & \textbf{pa} & 1.5K & 24.2K & 57.0K \\
\textbf{hi} & 2.2K & 39.7K & 79.7K & \textbf{sa} & 1.8K & 19.0K & 41.6K \\
\textbf{kn} & 1.2K & 12.8K & 25.7K & \textbf{sat} & 2.9K & 34.8K & 83.4K \\
\textbf{ks} & 0.7K & 7.3K & 22.2K & \textbf{sd} & 2.2K & 33.6K & 94.0K \\
\textbf{kok} & 0.8K & 8.3K & 23.0K & \textbf{ta} & 2.7K & 31.5K & 97.4K \\
\textbf{mai} & 2.5K & 33.1K & 90.7K & \textbf{te} & 2.6K & 29.1K & 92.7K \\
\textbf{ml} & 1.2K & 13.7K & 24.9K & \textbf{ur} & 0.2K & 3.4K & 4.6K \\
\bottomrule
\end{tabular}

\end{table}
\endgroup

\begingroup
\setlength{\tabcolsep}{2.8pt} 
\begin{table*}[ht!]
\centering
\caption{Comparison of ASR systems on WER and OIWER. The intensity of the green indicates difference between WER and OIWER. Due to space constraints, we present results for the top 11 languages and report the average for the remaining 11 in the `others' column. }
\label{tab:main_results}
\scriptsize

\begin{tabular}{c|cc|cc|cc|cc|cc|cc|cc|cc|cc|cc|cc|cc}
\toprule
\multicolumn{1}{l|}{} & \multicolumn{2}{c|}{\textbf{bn}} & \multicolumn{2}{c|}{\textbf{gu}} & \multicolumn{2}{c|}{\textbf{hi}} & \multicolumn{2}{c|}{\textbf{kn}} & \multicolumn{2}{c|}{\textbf{ml}} & \multicolumn{2}{c|}{\textbf{mr}} & \multicolumn{2}{c|}{\textbf{or}} & \multicolumn{2}{c|}{\textbf{pa}} & \multicolumn{2}{c|}{\textbf{ta}} & \multicolumn{2}{c|}{\textbf{te}} & \multicolumn{2}{c|}{\textbf{ur}} & \multicolumn{2}{c}{\textbf{others (avg.)}} \\
 & wer & oiwer & wer & oiwer & wer & oiwer & wer & oiwer & wer & oiwer & wer & oiwer & wer & oiwer & wer & oiwer & wer & oiwer & wer & oiwer & wer & oiwer & wer & oiwer \\
\midrule
{\raisebox{-0.2em}{\includegraphics[height=1.em]{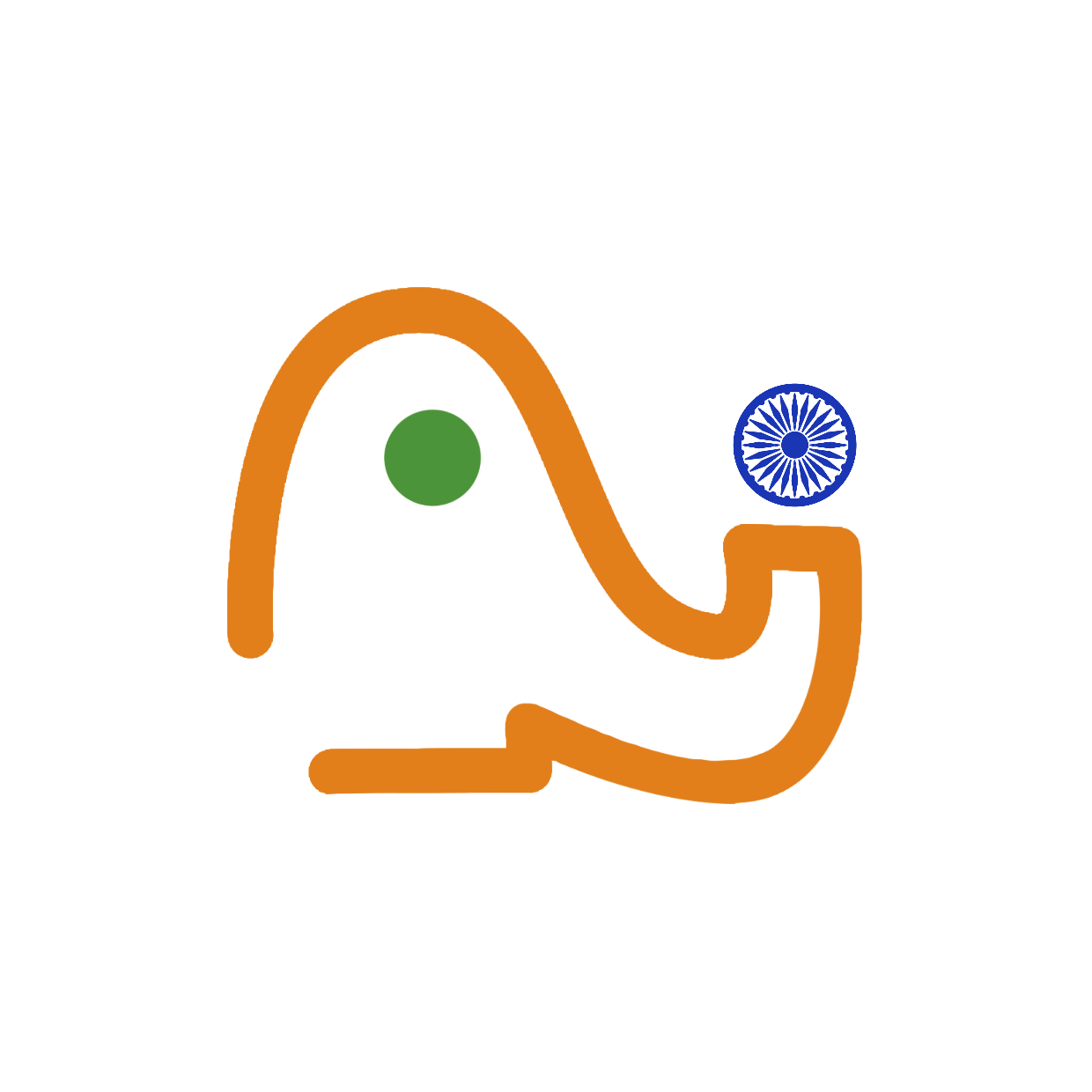}}}-C & 11.2 & \cellcolor[HTML]{E6F5EE}7.7 & 14.6 & \cellcolor[HTML]{DCF1E6}9.7 & 13.2 & \cellcolor[HTML]{E0F3EA}8.9 & 24.3 & \cellcolor[HTML]{C7E9D8}16.6 & 30.1 & \cellcolor[HTML]{98D6B7}16.0 & 11.7 & \cellcolor[HTML]{E5F5ED}8.1 & 19.7 & \cellcolor[HTML]{C3E7D5}11.4 & 10.7 & \cellcolor[HTML]{EBF7F1}7.9 & 27.0 & \cellcolor[HTML]{B8E2CD}17.2 & 22.9 & \cellcolor[HTML]{BCE4D0}13.7 & 5.4 & \cellcolor[HTML]{EEF8F3}3.0 & 23.2 & \cellcolor[HTML]{D2EDE0}17.0 \\
{\raisebox{-0.2em}{\includegraphics[height=1.em]{ai4b.png}}}-IC & 13.4 & \cellcolor[HTML]{E4F4EC}9.6 & 16.9 & \cellcolor[HTML]{D9F0E4}11.6 & 14.8 & \cellcolor[HTML]{DEF2E8}10.2 & 27.0 & \cellcolor[HTML]{C4E7D6}18.9 & 32.2 & \cellcolor[HTML]{9DD8BB}18.8 & 14.2 & \cellcolor[HTML]{E3F4EC}10.3 & 20.6 & \cellcolor[HTML]{C1E6D4}12.1 & 12.5 & \cellcolor[HTML]{ECF7F2}9.8 & 30.1 & \cellcolor[HTML]{B4E1CB}19.8 & 25.5 & \cellcolor[HTML]{BDE5D2}16.5 & 6.4 & \cellcolor[HTML]{EDF8F3}3.9 & 23.5 & \cellcolor[HTML]{D5EEE2}17.7 \\
{\raisebox{-0.2em}{\includegraphics[height=1.em]{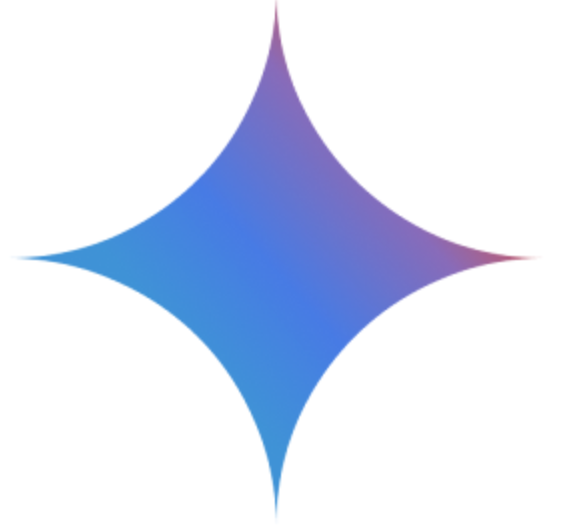}}} & 22.2 & \cellcolor[HTML]{BBE4CF}12.8 & 18.6 & \cellcolor[HTML]{D1EDDF}12.3 & 19.5 & \cellcolor[HTML]{C6E8D7}11.6 & 40.9 & \cellcolor[HTML]{9BD7BA}27.2 & 37.0 & \cellcolor[HTML]{89CFAD}20.8 & 19.1 & \cellcolor[HTML]{D1EDDF}12.8 & 30.3 & \cellcolor[HTML]{98D6B7}16.2 & 21.2 & \cellcolor[HTML]{C1E6D4}12.7 & 44.4 & \cellcolor[HTML]{57BB8A}21.5 & 33.4 & \cellcolor[HTML]{99D6B8}19.4 & 10.2 & \cellcolor[HTML]{E3F4EC}6.3 & 49.7 & \cellcolor[HTML]{94D4B5}35.1 \\
{\raisebox{-0.2em}{\includegraphics[height=1.em]{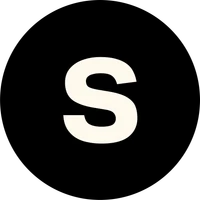}}} & 25.2 & \cellcolor[HTML]{ACDEC5}13.8 & 23.3 & \cellcolor[HTML]{D4EEE1}17.4 & 19.9 & \cellcolor[HTML]{CFECDE}13.3 & 37.0 & \cellcolor[HTML]{9BD7BA}23.3 & 40.3 & \cellcolor[HTML]{88CFAC}24.0 & 21.3 & \cellcolor[HTML]{D2EDE0}15.1 & 31.9 & \cellcolor[HTML]{A3DABF}19.3 & 21.8 & \cellcolor[HTML]{CEECDD}15.1 & 44.1 & \cellcolor[HTML]{68C296}23.5 & 33.5 & \cellcolor[HTML]{A1D9BE}20.6 & - & - & - & - \\
{\raisebox{-0.2em}{\includegraphics[height=1.em]{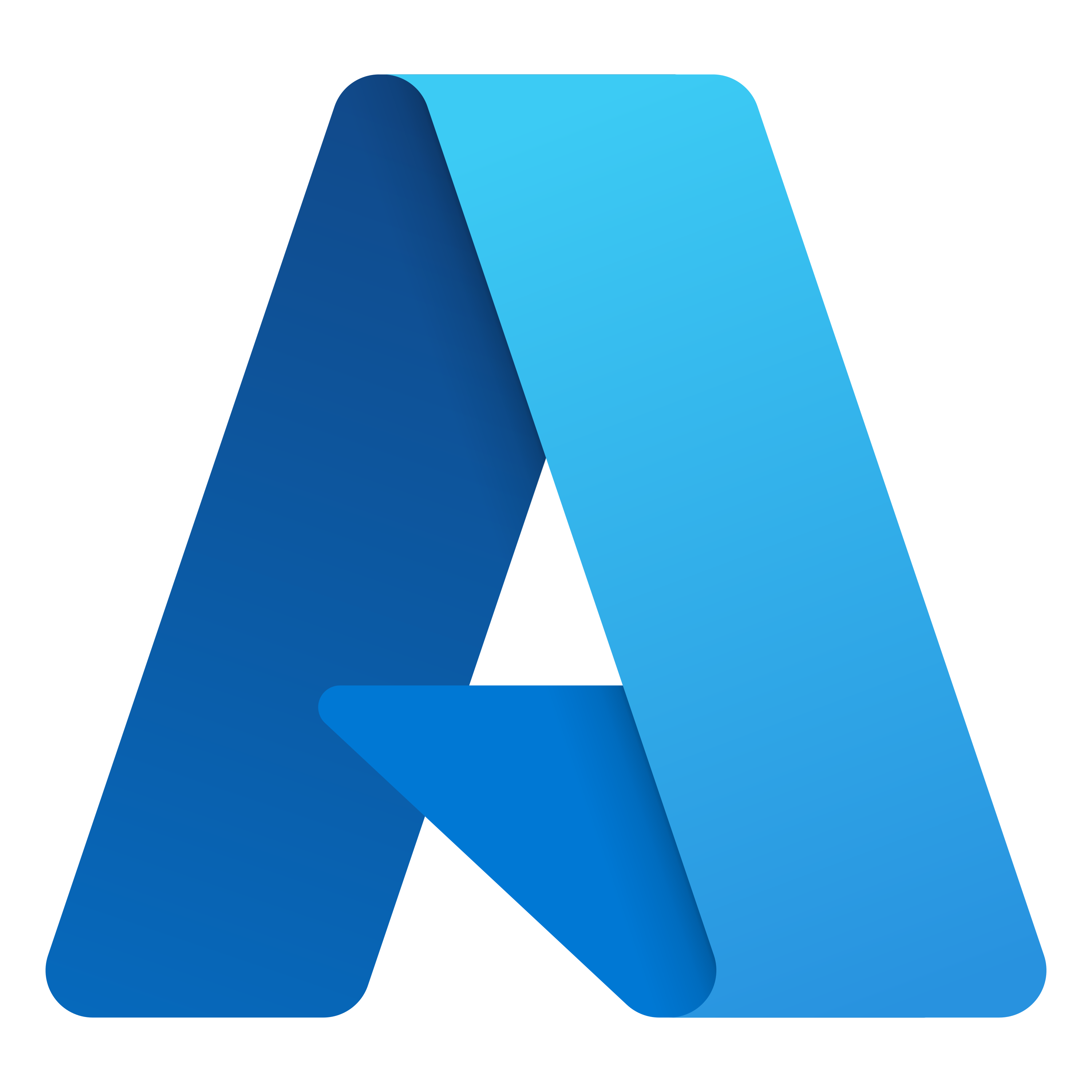}}} & 34.8 & \cellcolor[HTML]{9AD7B9}21.0 & 49.1 & \cellcolor[HTML]{D9F0E4}43.8 & 22.2 & \cellcolor[HTML]{CEEBDD}15.4 & 53.4 & \cellcolor[HTML]{B8E2CD}43.6 & 53.5 & \cellcolor[HTML]{87CFAC}37.1 & 35.9 & \cellcolor[HTML]{CFECDE}29.3 & 38.1 & \cellcolor[HTML]{A5DBC0}25.7 & 38.9 & \cellcolor[HTML]{CAEADA}31.6 & 61.4 & \cellcolor[HTML]{7DCBA5}43.6 & 52.4 & \cellcolor[HTML]{A8DCC3}40.5 & 18.1 & \cellcolor[HTML]{DAF0E5}13.0 & - & - \\
{\raisebox{-0.2em}{\includegraphics[height=1.em]{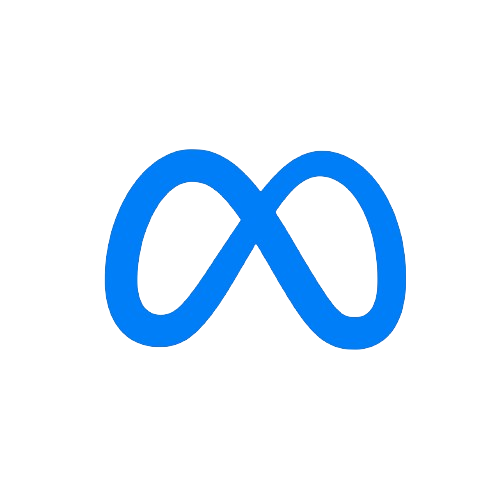}}} & 47.2 & \cellcolor[HTML]{D0ECDE}40.7 & 48.3 & \cellcolor[HTML]{DDF2E7}43.6 & 43.2 & \cellcolor[HTML]{DFF2E9}38.8 & 67.6 & \cellcolor[HTML]{DBF1E6}62.6 & 68.5 & \cellcolor[HTML]{CEEBDD}61.7 & 53.6 & \cellcolor[HTML]{DBF1E6}48.6 & 54.7 & \cellcolor[HTML]{B9E3CE}45.1 & 54.3 & \cellcolor[HTML]{D8EFE4}48.9 & 79.1 & \cellcolor[HTML]{DAF0E5}74.0 & 68.5 & \cellcolor[HTML]{D0ECDE}62.0 & 26.5 & \cellcolor[HTML]{E4F4EC}22.7 & - & - \\

\bottomrule
\end{tabular}

\end{table*}
\endgroup

\noindent\textbf{ASR Systems Compared.} We consider the following ASR systems in our study, covering both open source models and publicly available APIs for Indian languages.

\noindent \textit{Canary} \cite{canary}  {\raisebox{-0.4em}{\includegraphics[height=1.5em]{ai4b.png}}}-C: We finetune the Canary-1B-flash \cite{canary} model on pseudo-labeled transcripts of the MahaDhwani \cite{indicconformer} dataset consisting of 200K hours of data across 22 Indian languages.

\noindent \textit{IndicConformer} \cite{indicconformer} {\raisebox{-0.4em}{\includegraphics[height=1.5em]{ai4b.png}}}-IC: an open-source multilingual Conformer model released by AI4Bharat with support for 22 Indian languages.

\noindent \textit{Saarika:v2} {\raisebox{-0.2em}{\includegraphics[height=1.0em]{sarvam.png}}}: a publicly available Speech-To-Text API released by SarvamAI, supporting 10 Indian languages.

\noindent \textit{Meta MMS-1B-all} \cite{mms} {\raisebox{-0.45em}{\includegraphics[height=1.5em]{meta.png}}}: an open-source massively multilingual speech model from Meta supporting 
14 Indian languages. 

\noindent \textit{Azure}~{\raisebox{-0.2em}{\includegraphics[height=1.em]{azure.png}}}: a batch transcription API (version:2024-06-01) provided by Microsoft Azure with support for 10 Indian languages.

\noindent \textit{Gemini-2.5-Pro} \cite{gemini} {\raisebox{-0.35em}{\includegraphics[height=1.2em]{gemini.png}}}: Google's Gemini-2.5-Pro model used 
by passing the following prompt ``Transcribe the audio in the $<$lang$>$ language'', where $<$lang$>$ is the name of the language.


\section{Results and Discussion}

\subsection{Evaluating ASR models using OIWER}
Table \ref{tab:main_results} shows that the overall OIWER is lower than WER, across all models and languages. This is expected and aligns with both our perception of model quality as ASR practitioners and the feedback from downstream users regarding perceived quality as shown in Figure \ref{fig:motivation}. Next, we observe that when comparing models, WER indicates that Gemini is 18.1 points worse than Canary on average. However, OIWER suggests a smaller difference of 11.5 points, offering a more accurate representation of model performance. This is particularly evident in Tamil, where the WER difference of 17.4 drops to 4.3 when using OIWER. This suggests that WER may exaggerate performance gaps, potentially providing a false sense of progress. Azure’s model shows a similar trend, confirming that OIWER mitigates over-penalization compared to standard WER.

Additionally, we observe differences in language-specific evaluation. For Malayalam and Tamil, the difference between WER and OIWER is 13.9 and 14.4, respectively, when averaged over all models. This suggests variations impact these languages more. On the other hand, for languages with better orthographic standardization, such as Gujarati and Marathi, the average difference is smaller, at 5.4 and 5.3, respectively. We also observe that while Hindi has standardized spellings, its average difference is 5.8, attributed to the large number of loaned words, causing more variations.

\subsection{Does OIWER better reflect human-perceived WER?}
\begin{figure}
    \centering
    \includegraphics[width=0.7\linewidth]{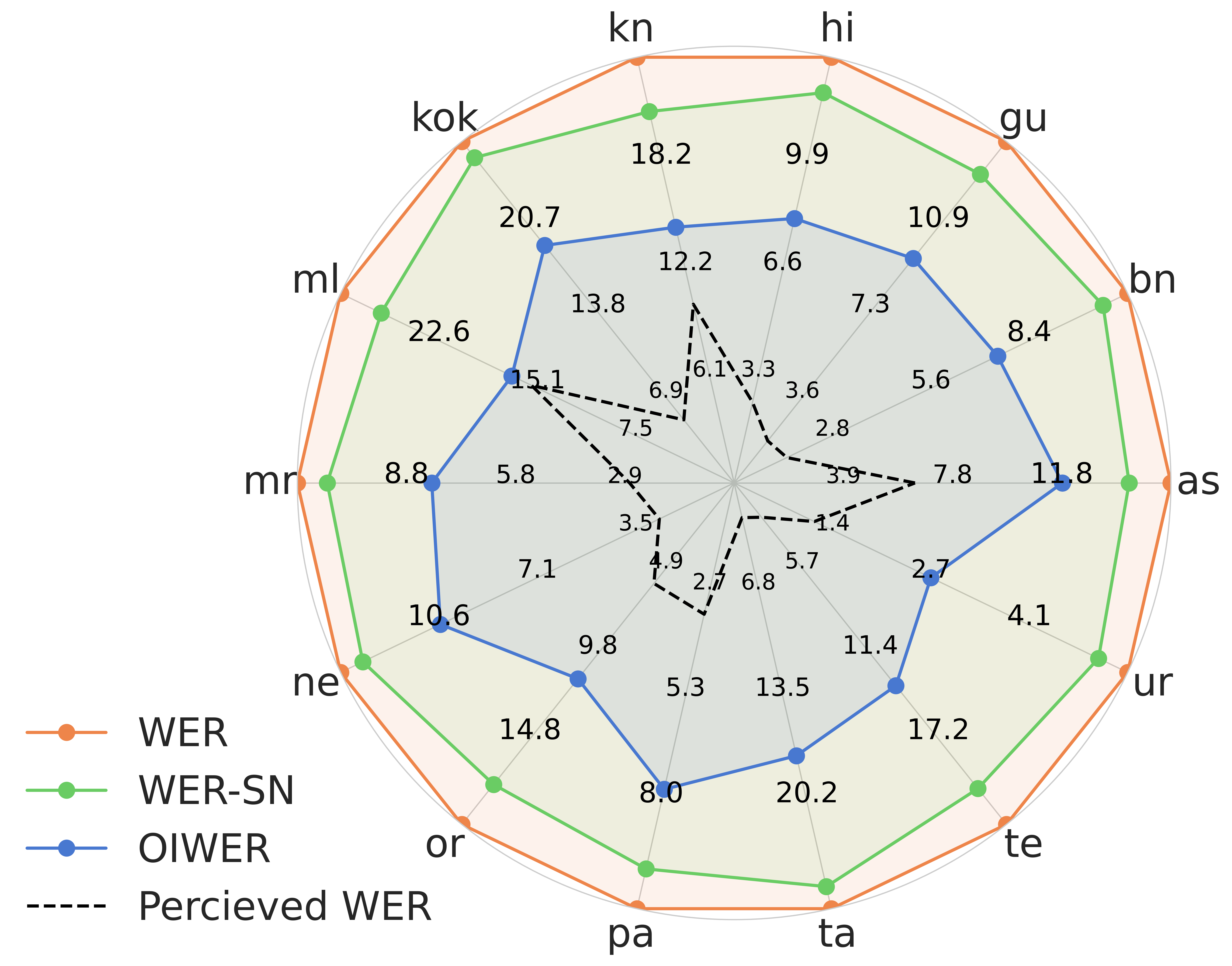}
    \caption{OIWER best aligns with perceived WER across languages.}
    \label{fig:baseline_comparison}
\end{figure}

We compare our approach to \cite{snwer}, which introduces Spelling and Segmentation Normalization (WER-SN) to address orthographic variations in Indian languages. While their work relies on an in-house grapheme-to-phoneme (G2P) system, we adopt the open-source eSpeak-ng for reproducibility. Figure \ref{fig:baseline_comparison} presents results across WER, WER-SN, and OIWER for Canary ({\raisebox{-0.4em}{\includegraphics[height=1.5em]{ai4b.png}}}-C). WER-SN reduces spelling variation penalties, yielding a 1.5-point average improvement over standard WER.
However, OIWER provides a more accurate performance score, closing the gap with perceived WER by an average of 4.9 points compared to WER-SN. The (human-)perceived WER is computed by asking humans to post-edit the model's output and then comparing this post-edited transcript with the model's output (as mentioned in section \ref{sec:motivation}). Further analysis revealed that WER-SN failed to capture spelling variations for loanwords and missed several colloquial word forms. This shows that while normalization-based methods capture some plausible variations, OIWER more fully represents the orthographic diversity of Indian languages.
Ideally, OIWER and perceived WER should be close. We indeed observe this for Malayalam where the gap is 1.5 WER. However, on average across languages, a gap of 6.9 WER remains, particularly high for Telugu and Konkani. Upon inspection, we find that this gap is due to inherent ambiguities in the audio that cannot be resolved without further context (one of the three factors mentioned in section \ref{sec:motivation}).

\subsection{Do variations reduce substitution errors?}
We compute the total number of substitution, insertion, and deletion errors for both WER and OIWER across all languages as shown in Figure \ref{fig:wer_scatter}(a). Our analysis shows that while the number of deletions and insertions remain similar, substitutions decrease significantly in OIWER. Specifically, the introduction of variations led to a reduction of 28.5K substitutions, summed across all languages. This reduction highlights that incorporating variations reduces false substitutions, validating the effectiveness of our approach.




\subsection{Can LLM-generated variations be used as a proxy for human-corrected variations?}
\begin{figure}
    \centering
    \includegraphics[width=1\linewidth]{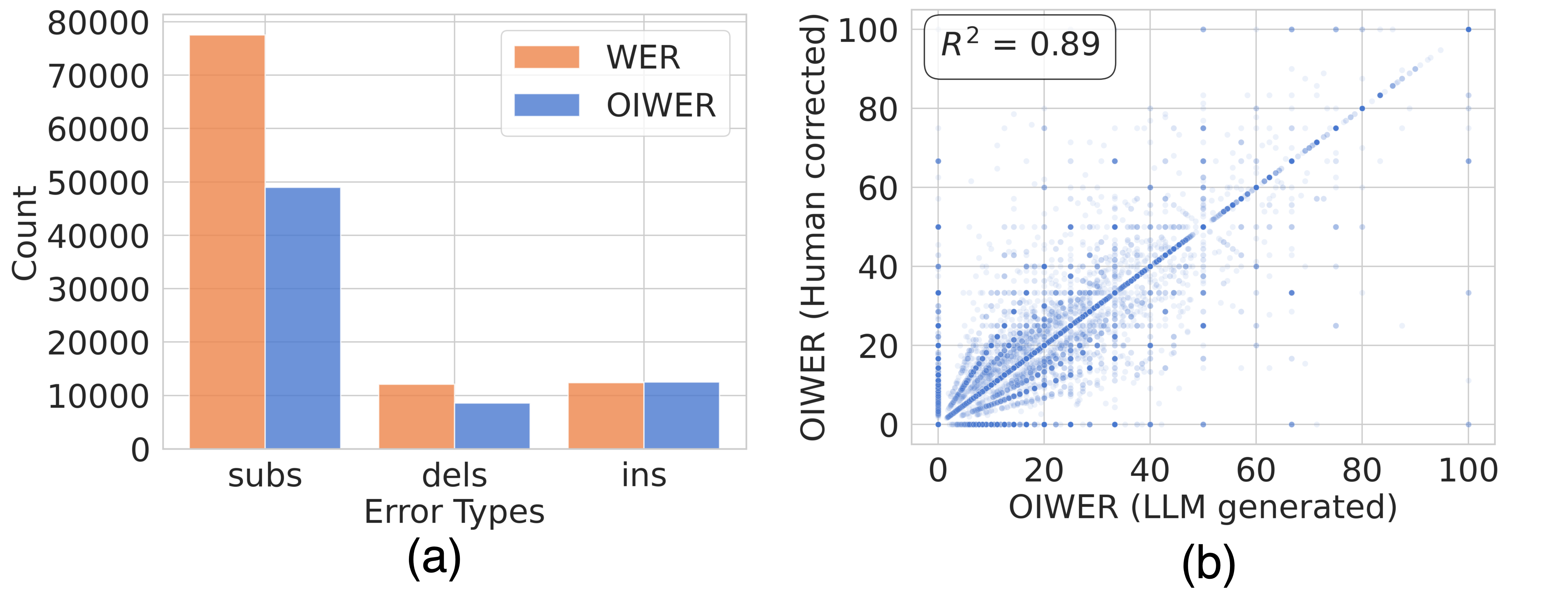}
    \caption{(a) OIWER eliminates false substitutions. (b) OIWER with LLM-generated variations shows a strong correlation with human-corrected variations.}
    \label{fig:wer_scatter}
\end{figure}

We created two orthographically-informed variations of the IndicVoices benchmark: one using variations suggested solely by the LLM (OIWER LLM), and the other using variations verified by human transcribers (OIWER Human). We then compute OIWER for each utterance in both benchmarks to investigate whether we can eliminate reliance on manual intervention, reducing time and cost, by using the LLM-generated variations instead. This analysis covers all 40.3K utterances across all languages.  The scatter plot in Figure \ref{fig:wer_scatter}(b) shows a strong correlation between OIWER values in the two benchmarks, with a coefficient of determination of 0.89. This suggests that LLM-generated variations can serve as a reliable proxy for human-corrected variations.

\section{Conclusion}
We present a comprehensive framework for orthographically informed (OI) evaluation of ASR systems for Indian languages that account for the full spectrum of orthographic variations, as verified by language experts. We introduce Orthographically-Informed Word Error Rate (OIWER) to account for these variations, offering a more robust and perceptually accurate measure of model performance. Our analysis highlights several key findings: (i) OIWER consistently yields lower scores than WER across all models and languages, providing a more realistic evaluation of ASR models, (ii) OIWER reduces over-penalization from substitution errors and aligns better with human perception, (iii) LLM-generated variations can serve as a reliable proxy for human-corrected variations, saving time and cost while maintaining evaluation quality.


\bibliographystyle{IEEEbib}
\bibliography{strings,refs}

\end{document}